# An End-to-end Entangled Segmentation and Classification Convolutional Neural Network for Periodontitis Stage Grading from Periapical Radiographic Images

Tanjida Kabir∗, Chun-Teh Lee†, Jiman Nelson†, Sally Sheng†, Hsiu-Wan Meng†, Luyao Chen∗,
Muhammad F Walji†, Xioaqian Jiang∗, and Shayan Shams‡

∗School of Biomedical Informatics, The University of Texas Health Science Center at Houston, Houston, Texas, USA

†Department of Periodontics and Dental Hygiene, The University of Texas Health Science Center at Houston, Houston, Texas, USA
‡Department of Applied Data Science, San Jose State University, San Jose, California, USA
Email: tanjida.kabir@uth.tmc.edu, Chun-Teh.Lee@uth.tmc.edu, Jiman.Nelson@uth.tmc.edu, Sally.Sheng@uth.tmc.edu,
Hsiu-Wan.Meng@uth.tmc.edu, Luyao.Chen@uth.tmc.edu, Muhammad.F.Walji@uth.tmc.edu, Xiaoqian.Jiang@uth.tmc.edu,
Shayan.Shams@sjsu.edu

*Abstract*—Periodontitis is a biofilm-related chronic inflammatory disease characterized by gingivitis and bone loss in the teeth area. Approximately 61 million adults over 30 suffer from periodontitis (42.2%), with 7.8% having severe periodontitis in the United States. The measurement of radiographic bone loss (RBL) is necessary to make a correct periodontal diagnosis, especially if the comprehensive and longitudinal periodontal mapping is unavailable. However, doctors can interpret X-rays differently depending on their experience and knowledge. Computerized diagnosis support for doctors sheds light on making the diagnosis with high accuracy and consistency and drawing up an appropriate treatment plan for preventing or controlling periodontitis. We developed an end-to-end deep learning network HYNETS (Hybrid NETwork for pEriodoNTiTiS STagES from radiograpH) by integrating segmentation and classification tasks for grading periodontitis from periapical radiographic images. HYNETS leverages a multi-task learning strategy by combining a set of segmentation networks and a classification network to provide an end-to-end interpretable solution and highly accurate and consistent results. HYNETS achieved the average dice coefficient of 0.96 and 0.94 for the bone area and tooth segmentation and the average AUC of 0.97 for periodontitis stage assignment. Additionally, conventional image processing techniques provide RBL measurements and build transparency and trust in the model's prediction. HYNETS will potentially transform clinical diagnosis from a manual time-consuming, and error-prone task to an efficient and automated periodontitis stage assignment based on periapical radiographic images.

## I. INTRODUCTION

Periodontitis is a biofilm-related chronic inflammatory disease characterized by gingivitis and bone loss in the area of the teeth [1]. According to the 2009-2014 National Health and Nutrition Examination Survey (NHANES), around 42.2% (61 million) adults over 30 years old in the United States have periodontitis, with 7.8% having severe periodontitis [2]. Periodontitis results in alveolar bone loss, edentulism, masticatory dysfunction, gingivitis, and chewing disorders leading to malnutrition [3]. Identifying the stage of periodontitis, which represents the severity of the disease, is necessary to make a correct periodontal diagnosis, especially if the comprehensive and longitudinal periodontal mapping is not available [4]. However, depending on individual experience, clinical examination, and X-ray parameters, different interpretations of X-ray images may occur, highlighting the requirement of Computer-Aided Diagnosis (CAD) software to provide an accurate and consistent diagnosis.

Innovative screening plan and customized treatment programs are critical for complex personalized dental care to improve individual outcomes. Modern high-throughput omics methods generate high-dimensional data, but the resulting wealth of information is difficult to evaluate using traditional statistical methods [5] due to the requirement of feature selection or interaction specifications [6]. Specifying interactions is a complex problem since we do not know exactly which interactions contribute to the development of periodontitis. Therefore, the diagnosis and classification of diseases using machine learning algorithms has attracted a lot of attention [7]. Deep learning models have been extensively applied to various medical domains [8]–[12]. However, the development of deep learning-based CAD in oral imaging has been minimal. Lee et al. developed a Convolutional Neural Network (CNN) model to detect cephalometric landmarks from radiographic images [13]. Murata et al. developed a CNN model to classify maxillary sinusitis from panoramic radiography [14]. Deep learning models have also been utilized for tooth classification from dental radiographs [15], [16] and Cone-Beam Computed Tomography (CBCT) [17], carries detection from radiographic images [18]–[22], and apical lesions from panoramic dental radiographs [23]. Alalharith et al. developed Faster Region-based Convolutional Neural Networks to detect early signs of gingivitis in orthodontic patients [24]. Ezhov et al. developed a deep learning model to identify anatomical landmarks, pathologies, and clinical effectiveness and safety of their model using CBCT imaging [25]. You et al. detected plaque on primary teeth using a deep CNN model [26]. Kouznetsova et al. developed machine-learning models to distinguish oral cancer and periodontitis, using patients' saliva metabolites [27]. Liang et al. designed an interactive user app using a deep learning model for early diagnosis of five oral conditions such as caries, periodontitis, soft deposit, dental calculus, and dental discoloration [28].

Few studies have investigated the utilization of machine learning in periodontitis detection. Farhadian et al. developed a machine learning model using support vector machine (SVM) to diagnose different periodontal disease like Gingivitis, localized periodontitis, and generalized periodontitis [7]. Lee et al. developed a deep learning model to classify periodontally compromised teeth [29] from periapical radiographic images. A couple of studies utilized deep learning models to detect periodontal bone loss from panoramic dental radiographs [30], [31]. However, these studies only detect bone loss and do not assign a periodontitis stage to each tooth. Chang et al. developed a hybrid model using image processing and deep learning to assign periodontitis stage from panoramic radiographic images [32]. However, compared to periapical images, panoramic images generally have more bone level variation, lower interpretation agreement among clinicians, and more image enlargement [33]–[35] and therefore, periapical images are considered the "gold standard" for periodontal diagnosis. However, measuring Radiographic Bone Loss (RBL) and assigning periodontitis stages from periapical images majorly depends on individual experience and knowledge since it requires correct identification of bone area, tooth, and the cemento-enamel junction (CEJ) line for each tooth in the image. To the best of our knowledge, we are the first team to develop a deep learning-based CAD tool to measure bone loss and assign a periodontitis stage for each tooth from periapical radiographic images.

Our main contributions can be divided into three parts: (1) methodology: We have developed HYNETS (Hybrid NETwork for pEriodoNTiTiS STagES from radiograpH) by integrating several segmentation networks and a classification network through a multi-task learning strategy to take into account the discrepancies between segmentation annotators and clinical examiners and provide a consistently interpretable solution for periodontal diagnosis and high-precision results; (2) interpretability: While our implementation uses a classification model to assign the final stage of bone loss to each tooth, we use image analysis techniques and rules-based methods to provide better interpretation and insight into the deep learning model that can help physicians identify phase assignment glitches and potential errors caused by segmentation affecting phase assignment. (3) Performance: Our model provides a more accurate assessment of bone level for the diagnosis of periodontitis than all previously reported models.

## II. METHODS

### A. Model overview

Figure 1 illustrates high level architecture of HYNETS. The model is designed specifically to provide high classification accuracy while maintaining interpretability.

As it is illustrated in Figure 1, the segmentation networks generate bone area, tooth, and the CEJ line masks from the periapical radiograph images. These masks then go through series of postprocessing steps for noise removal and quality improvement. Next, the masks are overlayed, and the RBL percentage is calculated from the overlayed mask, Figure 1(a) using image analysis techniques as an intermediate output for interpretability. Finally, the input of the classification network is obtained by extracting each tooth and its corresponding CEJ line and bone area, Figure 1(b), to predict the periodontitis stage for each tooth.

In this work, the periodontitis stage assignment based on RBL for each tooth and periodontal diagnosis of the whole dentition is based on the 2018 periodontitis classification [4] defining stage I: RBL<15% (in the coronal third of the root), stage II: 15%<RBL<33% (in the coronal third of the root), stage III: extending to the middle third of root and beyond (RBL>33%).

### B. Segmentation task

We evaluated variants of U-Net [36] architecture to find a solution with the highest accuracy for the bone area, tooth, and CEJ line segmentation. We compared the performance of using a single segmentation network for the bone area, tooth, and CEJ line segmentation with three separate networks. The average DSC and Jaccard index (JI) for three different networks is (DSC = 0.94, JI = 0.91) while a single network with three channels only achieved (DSC = 0.72, JI = 0.62) due to the lack of shared inductive biases among tasks. This is additionally supported by the comparisons in Table I, showing that no single size fits all and customized network structures and hyperparameters have performance advantages over a single model.

We additionally evaluated multiple variants of U-Net and observed U-Net with ResNet-34 [37] as the encoder, Figure 2(a), performs the best for bone area and teeth segmentation. Replacing the CNN encoder with ResNet-34 enables the model to learn more features by having a deeper network while elevating vanishing gradient problems using shortcut paths. U-Net with CNN encoder, Figure 2(b), outperformed other variants for the CEJ line detection. The evaluation results with different architecture are shown in Table I.

As it is shown in the Figure 2(a,b), the U-Net architecture consists of an encoder path (left path) including convolutional layers (ResNet block for ResNet encoder) and decoder path (right path), which includes up-sampling and convolutional layers. In every step, the decoder up-samples the feature map using the nearest neighbor algorithm and performs a 2 × 2 convolution on the up-sampled feature map. Finally, the output is concatenated with the correspondingly cropped feature map from the encoder path using cross-connection between encoder and decoder, fusing extracted feature maps of each resolution in the encoding path with different scales in the decoding path.

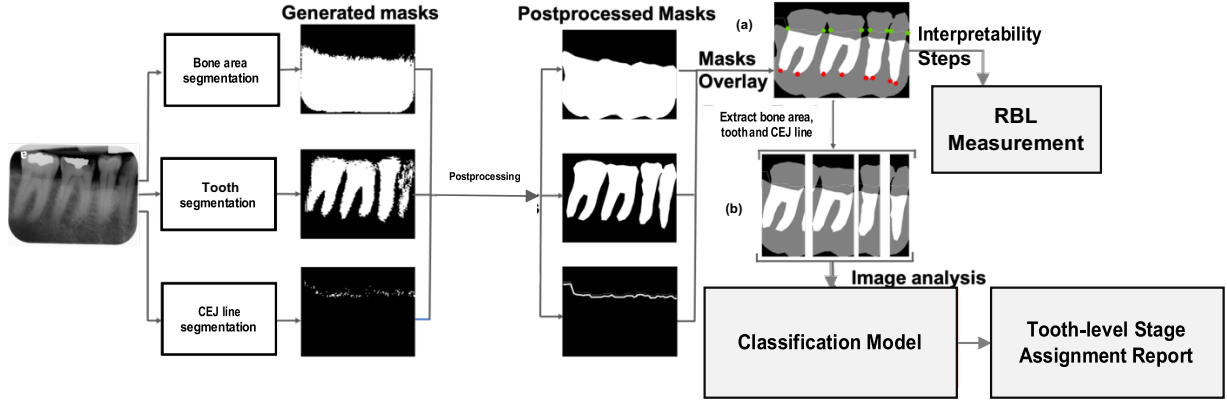

**Fig. 1:** Flow diagram of the HYNETS.

**TABLE I:** Performance evaluation of different model variants for bone area, tooth and the CEJ line segmentation.

| Task Name | Model | Dice Coefficient | Jaccard Index | Pixel Accuracy |
|---|---|---|---|---|
| Bone Area Segmentation | U-Net | 0.9621 | 0.9319 | 0.9570 |
|  | U-Net with ResNet-50 Encoder | 0.9229 | 0.9333 | 0.9603 |
|  | **U-Net with ResNet-34 Encoder** | **0.9635** | **0.9343** | **0.9610** |
|  | U-Net with three channels | 0.9493 | 0.9064 | – |
| Tooth Segmentation | U-Net | 0.8994 | 0.8225 | 0.8413 |
|  | U-Net with ResNet-50 Encoder | 0.9533 | 0.9026 | 0.8660 |
|  | **U-Net with ResNet-34 Encoder** | **0.9470** | **0.9143** | **0.8898** |
|  | U-Net with three channels | 0.6150 | 0.4535 | – |
| CEJ Line Segmentation | U-Net with 3X3 kernel | 0.4287 | 0.2753 | 0.9456 |
|  | U-Net with 7X7 kernel | 0.6719 | 0.5143 | 0.9850 |
|  | **U-Net with 5X5 kernel** | **0.9129** | **0.8776** | **0.9966** |
|  | U-Net with three channels | 0.6205 | 0.5017 | – |

The binary cross-entropy loss is used as a loss function for the segmentation model. Equation 1 illustrates the formula to calculate binary-cross entropy loss.

$$\text{Segmentation Loss} = -\frac{1}{\|B\|} \sum_{I \in B} \sum_{i \in I} (y_i \log \hat{y}_i + (1-y_i) \log(1-\hat{y}_i)) \quad (1)$$

Here $\hat{y}_i$ is the value for a pixel in the prediction, $y_i$ is the target value of the pixel in the mask, $I$ is an image, and $B$ is the batch size.

### C. Postprocessing

We use Gaussian filters to reduce the noise of the generated masks. For the CEJ line, the model's output consists of a series of tightly arranged pixels that are not zero. We recognize the first unequal value from the left as the starting point and move one pixel to the right to find the nearest point of the line. We repeat this process until the end of the image to get the connected CEJ line.

Figure 1 displays the input images, segmentation models' generated masks and masks after postprocessing. Generated masks are overlayed to mark bone area, teeth, and the CEJ line (Figure 1(a)). Each tooth is extracted from the overlayed mask by drawing a bounding box surrounding each tooth using the generated teeth mask's coordinates. These cropped individual teeth with their corresponding CEJ line and bone area (1(b)) are inputs for the classification model for periodontitis stage assignment.

### D. Classification Task

We developed a fully convolutional networks to assign the periodontal stages to each tooth automatically. As it is illustrated in Figure 2(c), the model consists of four consecutive convolution layers followed by max-pooling. The output of the last max-pooling layer goes to a global average pooling, dropout, and two fully connected layers.

In the post-processing step, the classification network loops over the extracted teeth and predicts a periodontitis stage for each tooth in the input image. The categorical cross-entropy, shown in Equation 2 is used as the loss function.

$$Classification\ Loss = -\frac{1}{N}\sum_{i=1}^{N} y_i \log(f(s)_i) \quad (2)$$

Here $N$ is the batch size, $y_i$ is the true class for input $i$, and $f(s)_i$ is the model prediction. The output of this classification model is the final periodontal stage of each tooth in the image.

### E. Interpretability

To provide better insights into the HYNETS stage assignment, build trust in the model prediction, and avoid the "black box" behavior of deep learning models, we use image analysis methods to calculate RBL percentage from the intermediate products (overlayed mask) of the network to interpret the final classification result and provide reasoning for deep learning automatic stage assignment in an end-to-end manner.

The steps to calculate RBL percentage are as follows:

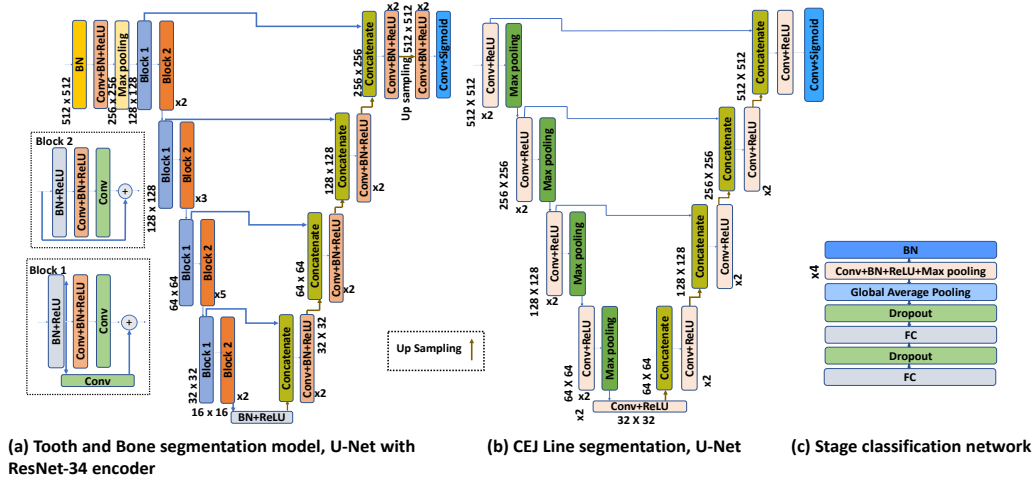

**Fig. 2:** High level architecture of the HYNETS.

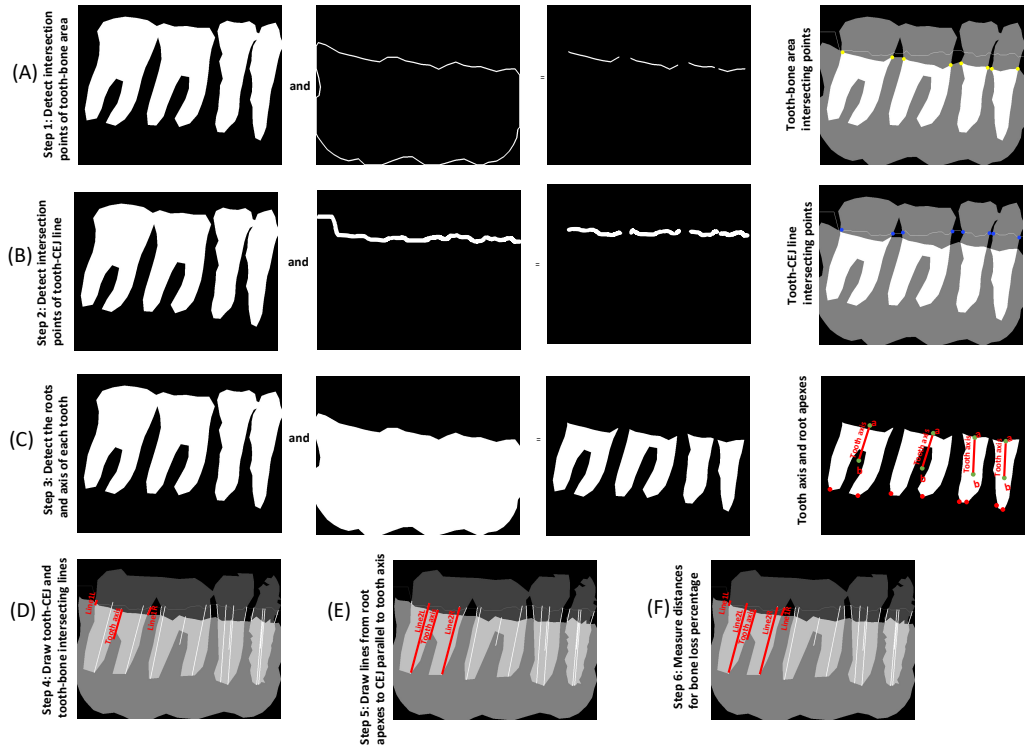

**Fig. 3:** Process of introducing interpretability and calculating bone loss percentage

*(i)* Perform logical "AND" operation between the bone area's contour in bone area mask and teeth mask to find the intersecting line of the teeth and bone area and mark the start and endpoint of the intersecting line (yellow dots in Figure 3a). *(ii)* Repeat the same process for the CEJ line and teeth mask to find the intersecting line of teeth and the CEJ line and mark the start and endpoint of the line (green dots in Figure 3b). *(iii)* Perform a logical "AND" operation between the teeth and bone area masks to detect the root and axis for each tooth. First, draw a contour surrounding each tooth and find the center points of these contours (point *b* in Figure 3c). Then, find the middle of intersecting line of each tooth and bone area (point a in Figure 3c). Next, draw a line crossing point a, b (the orange marked points) representing axis of the tooth. Finally, get the minimum point in the left and right side of the axis to get the roots' apexes (red dots in Figure 3c). *(iv)* Draw a line between the starting point of intersecting line between the tooth and CEJ line, and the starting point of intersecting line between tooth and bone area to get Line 1L. Repeat the same process for ending points to get Line 1R in Figure 3d. *(v)* Draw a line from each root apex to the CEJ line parallel to the tooth axis (Figure 3e Line 2L and Line 2R). *(vi)* Use equation 3 to obtain RBL percentage (Figure 3f).

TABLE II: Segmentation model accuracy for radiographic and heatmap images

|  | Bone Area Model | Tooth Model | CEJ Line Model |
|---|---|---|---|
| Radiographic images | 0.8924 | 0.9122 | 0.7544 |
| Heatmap images | 0.9610 | 0.8898 | 0.9966 |

TABLE III: DSC, Jaccard Index and Pixel Accuracy for segmentation model.

| Paper | Task | DSC | Jaccard Index | Pixel Accuracy |
|---|---|---|---|---|
| Chang et al. [32] | Bone Area Segmentation | 0.93 | 0.88 | 0.92 |
|  | Teeth Segmentation | 0.91 | 0.83 | 0.87 |
|  | CEJ Line Segmentation | 0.91 | 0.84 | 0.87 |
| HYNETS | Bone Area Segmentation | **0.9635** | **0.9343** | **0.9610** |
|  | Teeth Segmentation | **0.9470** | **0.9143** | **0.8898** |
|  | CEJ Line Segmentation | **0.9129** | **0.8776** | **0.9966** |

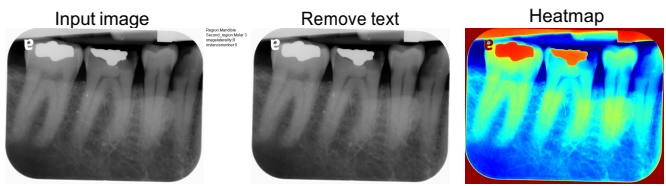

Fig. 4: Visualization of image preprocessing steps.

$$\text{Bone loss percentage} = \max\left(\frac{\text{Line 1L}}{\text{Line 2L}}, \frac{\text{Line 1R}}{\text{Line 2R}}\right) * 100 \quad (3)$$

*F. Dataset & Training*

Our dataset includes 700 periapical X-rays. All images are annotated and examined by three independent examiners, a board-certified periodontist professor, a board-certified clinical periodontist, and one resident in the periodontics program. Each tooth was assigned a periodontitis stage by all three examiners based on the 2018 periodontitis classification. The dataset was divided into 70%, 20%, and 10% for training, testing, and validation. The end-to-end HYNETS model was additionally tested on ten individual cases, with 10-12 x-rays per case. As already mentioned, since the assignment of the bone loss stage largely depends on the experience and knowledge of individuals, we use majority voting to obtain the ground truth for the final assignment if there was a conflict between the examiners.

In the data preprocessing step, the irrelevant texts from images are removed, and then heatmaps from those images are generated, shown in Figure 4. As it is shown in Table II, heatmap images increased the model performance because the variation of color intensity of heatmap images can give better visual cues for models to understand the features of images [8], [38]. In addition, image augmentation techniques such as cropping, flipping, rotating, and varying pixel intensities are used to increase the size and diversity of training data. Finally, the images are resized into $512 \times 512$ for training.

The segmented models are trained separately based on the annotated bone area, teeth, and the CEJ line. The classification model is trained by extracting each tooth from the images, using the examiners' annotations and their assigned stages as the ground truth. After a separate training session, all three segmentation networks and the classification network are fine-tuned. The total loss is an addition of the three segmentation losses for teeth, bone area, and CEJ line and classification loss. All parameters are updated to minimize the total loss in the training data. It is important to note that post-processing and tooth extraction are fused as automatic intermediate steps in our end-to-end training. Stochastic gradient descent with Adam optimizer with a learning rate of 0.001 and a batch size of 128 was used for separate training and fine-tuning the end-to-end model.

III. EXPERIMENTS & RESULTS

HYNETS integrates segmentation networks and a classification network, and therefore we evaluate its performance for both the segmentation task and classification task.

*A. Segmentation Task*

We reported Dice Similarity Coefficient (DSC), Jaccard Index, and Pixel Accuracy for our segmentation models in Table III. While to the best of our knowledge, HYNETS is the first DL-based CAD for bone loss stages from periapical radiographic images, we compare our segmentation result with the only other previous work performing segmentation on dental radiographs, and the comparison is shown in Table III. However, Chang et al. used panoramic radiographs while HYNETS was trained and tested with periapical radiographs. As it is shown in Table III, HYNETS consistently achieves higher segmentation performance metrics than the other work.

*B. Classification Task*

Figure 5 demonstrates the segmentation and classification models' performance to assign periodontal stages for each tooth and comparative results with expert's opinion.

We also show that HYNETS achieved average Area Under the Receiving Operating Characteristics Curve (AUC-ROC) of 0.96 for bone loss stage assignment by integrating segmentation and classification tasks using a multi-task learning technique. Figure 6a presents the AUC of bone loss stage assignment.

As illustrated in Table IV, HYNETS achieves higher classification AUC compared to previous studies and is the only model that is compatible with the 2018 periodontitis classification guideline working with periapical radiographs. However, we cannot directly compare the stage assignment accuracy with Chang et al. [32] since they used Mean Absolute Differences, which is not typical for a stage assignment (classification) task.

*C. RBL percentage measurement and stage assignment*

Figure 7 demonstrates the similarity of the RBL percentage for each individual tooth using Equation 3 with expert's opinion. Since HYNETS was trained with ground truth obtained from majority voting, to measure the amount of agreement between each expert and HYNETS in periodontitis stage assignment, we calculated Cohen's Kappa values between experts and HYNETS, presented in Table V, for ten additional cases (120 images) not included in training and test set. The HYNETS' classification result showed the highest agreement with the professor's ($\kappa$=0.6998, substantial agreement). Furthermore, the Kappa values highlight that periodontitis stage assignment depends on the examiner's knowledge and skills. The highest agreement among clinicians achieved between the professor and the clinical periodontist is substantial ($\kappa$=0.6265). On the other hand, the agreement decreases to fair ($\kappa$=0.4959) for the resident and professor.

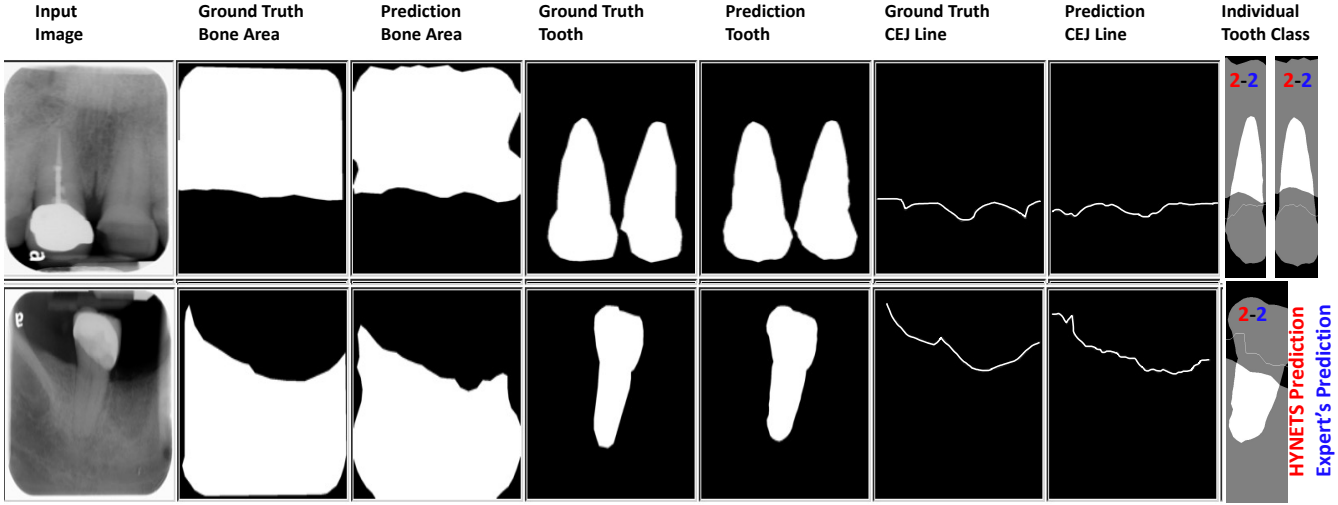

**Fig. 5:** Example of two radiographic images in different steps of HYNETS pipeline.

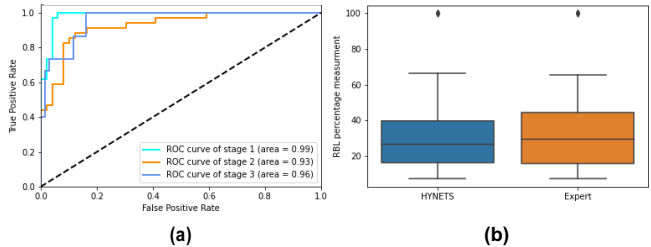

(a)         (b)

**Fig. 6:** (a) AUC achieved by HYNETS for each stage assignment. (b) Distribution of measured RBL by expert and HYNETS.

**TABLE IV:** Classification performance comparison with other works.

| Paper | Task | Classification | 2018 Guideline compatible | Image | AUC |
|---|---|---|---|---|---|
| Lee et al. [29] | Periodontally compromised tooth detection | Multiclass | X | Periapical | 0.826 |
| Krois et al. [30] | Bone Loss detection | Binary | X | Panoramic | 0.89 |
| Kim et al. [31] | Bone Loss detection | Binary | X | Panoramic | 0.92 |
| HYNETS | Bone Loss stage assignment for each tooth | Multiclass |  | Periapical | **0.96** |

We conducted a Student's t-test to identify a significant difference between the RBL percentage measured by experts and HYNETS, and the result showed no significant difference (p=0.42). Figure 6b illustrates the distribution of measured

**TABLE V:** The Cohen's Kappa values for experts and HYNETS.

|  | HYNETS | Professor | Clinical periodontist | Resident |
|---|---|---|---|---|
| HYNETS | 1.0 | **0.6998** | 0.4712 | 0.4959 |
| Professor | **0.6998** | 1.0 | 0.6265 | 0.4959 |
| Clinical periodontist | 0.4712 | 0.6265 | 1.0 | 0.3745 |
| Resident | 0.4959 | 0.4959 | 0.3745 | 1.0 |

RBL percentage for both expert and HYNETS, indicating no significant difference in the data distribution as well. HYNETS achieved high AUC in the periodontitis final stage assignment task while maintaining comparable measurement with the experts' highlighting the fact that HYNETS can provide high-quality segmentation and highly accurate classification results in an end-to-end manner.

### D. Clinical evaluation and reproducibility

We have also implemented a web interface to calculate the percentage of radiographic bone loss and assign periodontal stages to each tooth. An example screenshot of the web page for a periapical X-ray image is shown in Figure 8. Clinicians can upload a periapical X-ray image, obtain the corresponding bone loss percentage through an image analysis method, and get the highly accurate periodontal stage for each tooth using a classification model. The web interface also provides visualization of the bone area, tooth, and the CEJ line masks to provide better understanding and transparency for the whole process.

### IV. CONCLUSION

We introduced HYNETS (HybridNETwork for pEriodoN-TiTiS STagES from radiograpH), the first model to offer a fully interpretable deep learning solution for the assignment of periodontitis stages and RBL measurement from periapical X-rays. HYNETS uses a multi-task learning strategy by combining a set of segmentation networks and a classification network to deliver a consistently interpretable solution and highly accurate and consistent results.

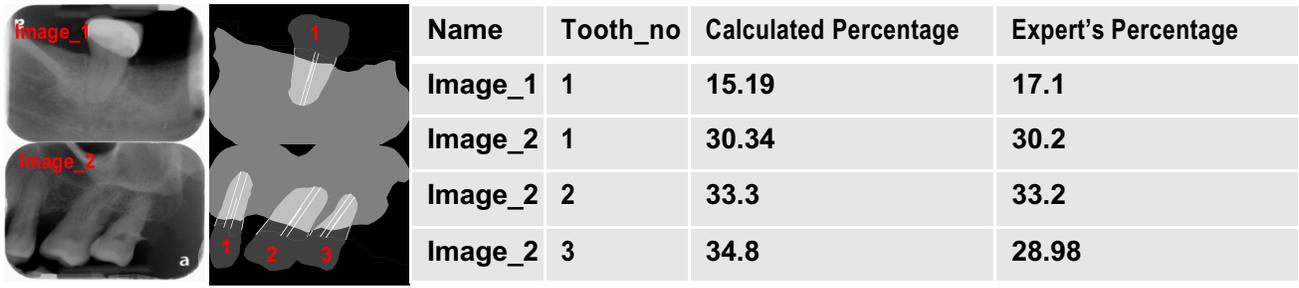

**Fig. 7:** Visual examples of the RBL percentage for two radiographic images and their corresponding expert's percentage.

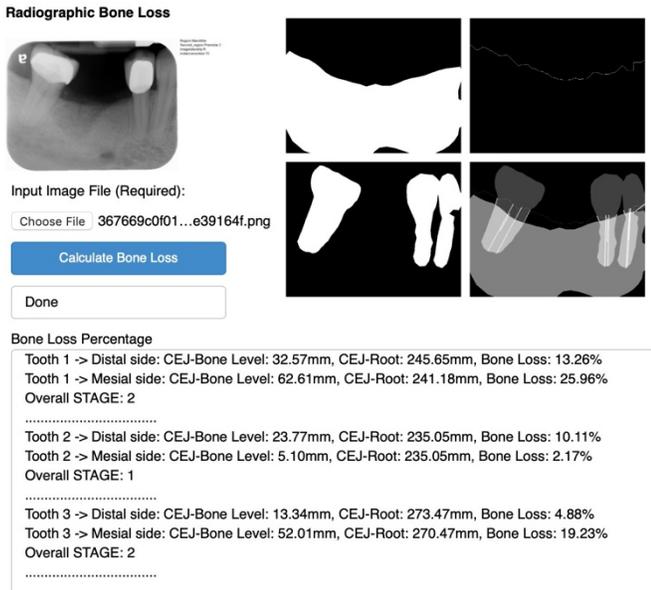

**Fig. 8:** Web interface for radiographic bone loss percentage and stage assignment.

Due to the fact that (i) the average DSC and JI among segmentation annotators is 0.85, 0.80, showing variability in the annotation; (ii) Kappa score table V indicates that human experts exhibit a large variability of agreements and do not fully agree with the initial segmentation annotation (otherwise, their rules-based judgments would produce the same staging); (iii) the imperfection of the segmentation networks leads to propagation of errors in the stage assignment, we design a robust framework by integrating segmentation and classification tasks to efficiently handle the errors caused by discrepancies among segmentation annotators and segmentation models. As a result, the average AUC for rule-based stage assignment, which uses only segmentation models (using the rules-based method), is 0.81. It is 0.96 for our proposed model with the integration of segmentation and classification tasks. This confirms our hypothesis that our model is more robust and error tolerant.

The result showed better segmentation and classification performance for HYNETS compared to previous works. Furthermore, the periodontitis classification showed considerable agreement with the most experienced examiner, and the percentage RBL measurement differences with the expert's measurement were not significant. Thus, we have demonstrated the effectiveness of multi-task learning in our implementation and its role in maintaining high performance for segmentation and classification tasks. This encourages its future integration into clinical practice.

### ACKNOWLEDGEMENT

XJ is CPRIT Scholar in Cancer Research (RR180012), and he was supported in part by Christopher Sarofim Family Professorship, UT Stars award, UTHealth startup, the National Institute of Health (NIH) under award number R01AG066749, R01GM114612 and U01TR002062, and the National Science Foundation (NSF) RAPID 2027790.

## REFERENCES

[1] T. E. Van Dyke, P. M. Bartold, and E. C. Reynolds, "The nexus between periodontal inflammation and dysbiosis," *Frontiers in immunology*, vol. 11, 2020.

[2] P. I. Eke, G. O. Thornton-Evans, L. Wei, W. S. Borgnakke, B. A. Dye, and R. J. Genco, "Periodontitis in us adults: National health and nutrition examination survey 2009-2014," *The Journal of the American Dental Association*, vol. 149, no. 7, pp. 576–588, 2018.

[3] M. S. Tonetti, S. Jepsen, L. Jin, and J. Otomo-Corgel, "Impact of the global burden of periodontal diseases on health, nutrition and wellbeing of mankind: A call for global action," *Journal of clinical periodontology*, vol. 44, no. 5, pp. 456–462, 2017.

[4] M. S. Tonetti, H. Greenwell, and K. S. Kornman, "Staging and grading of periodontitis: Framework and proposal of a new classification and case definition," *Journal of periodontology*, vol. 89, pp. S159–S172, 2018.

[5] M. Kebschull and P. N. Papapanou, "Exploring genome-wide expression profiles using machine learning techniques," in *Oral Biology*. Springer, 2017, pp. 347–364.

[6] B. J. Wolf, E. H. Slate, and E. G. Hill, "Ordinal logic regression: A classifier for discovering combinations of binary markers for ordinal outcomes," *Computational statistics & data analysis*, vol. 82, pp. 152–163, 2015.

[7] M. Farhadian, P. Shokouhi, and P. Torkzaban, "A decision support system based on support vector machine for diagnosis of periodontal disease," *BMC Research Notes*, vol. 13, no. 1, pp. 1–6, 2020.

[8] S. Shams, R. Platania, J. Zhang, J. Kim, K. Lee, and S.-J. Park, "Deep generative breast cancer screening and diagnosis," in *International Conference on Medical Image Computing and Computer-Assisted Intervention*. Springer, 2018, pp. 859–867.

[9] D. Riquelme and M. A. Akhloufi, "Deep learning for lung cancer nodules detection and classification in ct scans," *AI*, vol. 1, no. 1, pp. 28–67, 2020.

[10] L. Sun, S. Zhang, and L. Luo, "Tumor segmentation and survival prediction in glioma with deep learning," in *International MICCAI Brainlesion Workshop*. Springer, 2018, pp. 83–93.

[11] Y. Zhou, L. Xie, E. K. Fishman, and A. L. Yuille, "Deep supervision for pancreatic cyst segmentation in abdominal ct scans," in *International conference on medical image computing and computer-assisted intervention*. Springer, 2017, pp. 222–230.

[12] D. Rajan, D. Beymer, S. Abedin, and E. Dehghan, "Pi-pe: A pipeline for pulmonary embolism detection using sparsely annotated 3d ct images," in *Machine Learning for Health Workshop*. PMLR, 2020, pp. 220–232.

[13] H. Lee, M. Park, and J. Kim, "Cephalometric landmark detection in dental x-ray images using convolutional neural networks," in *Medical Imaging 2017: Computer-Aided Diagnosis*, vol. 10134. International Society for Optics and Photonics, 2017, p. 101341W.

[14] M. Murata, Y. Ariji, Y. Ohashi, T. Kawai, M. Fukuda, T. Funakoshi, Y. Kise, M. Nozawa, A. Katsumata, H. Fujita *et al.*, "Deep-learning classification using convolutional neural network for evaluation of maxillary sinusitis on panoramic radiography," *Oral radiology*, vol. 35, no. 3, pp. 301–307, 2019.

[15] O. Ronneberger, P. Fischer, and T. Brox, "Dental x-ray image segmentation using a u-shaped deep convolutional network," in *International Symposium on Biomedical Imaging*, 2015, pp. 1–13.

[16] G. Jader, J. Fontineli, M. Ruiz, K. Abdalla, M. Pithon, and L. Oliveira, "Deep instance segmentation of teeth in panoramic x-ray images," in *2018 31st SIBGRAPI Conference on Graphics, Patterns and Images (SIBGRAPI)*. IEEE, 2018, pp. 400–407.

[17] Y. Miki, C. Muramatsu, T. Hayashi, X. Zhou, T. Hara, A. Katsumata, and H. Fujita, "Classification of teeth in cone-beam ct using deep convolutional neural network," *Computers in biology and medicine*, vol. 80, pp. 24–29, 2017.

[18] R. B. Ali, R. Ejbali, and M. Zaied, "Detection and classification of dental caries in x-ray images using deep neural networks," in *International Conference on Software Engineering Advances (ICSEA)*, 2016, p. 236.

[19] M. M. Srivastava, P. Kumar, L. Pradhan, and S. Varadarajan, "Detection of tooth caries in bitewing radiographs using deep learning," *arXiv preprint arXiv:1711.07312*, 2017.

[20] J. Choi, H. Eun, and C. Kim, "Boosting proximal dental caries detection via combination of variational methods and convolutional neural network," *Journal of Signal Processing Systems*, vol. 90, no. 1, pp. 87–97, 2018.

[21] J.-H. Lee, D.-H. Kim, S.-N. Jeong, and S.-H. Choi, "Detection and diagnosis of dental caries using a deep learning-based convolutional neural network algorithm," *Journal of dentistry*, vol. 77, pp. 106–111, 2018.

[22] T. Hiraiwa, Y. Ariji, M. Fukuda, Y. Kise, K. Nakata, A. Katsumata, H. Fujita, and E. Ariji, "A deep-learning artificial intelligence system for assessment of root morphology of the mandibular first molar on panoramic radiography," *Dentomaxillofacial Radiology*, vol. 48, no. 3, p. 20180218, 2019.

[23] T. Ekert, J. Krois, L. Meinhold, K. Elhennawy, R. Emara, T. Golla, and F. Schwendicke, "Deep learning for the radiographic detection of apical lesions," *Journal of endodontics*, vol. 45, no. 7, pp. 917–922, 2019.

[24] D. M. Alalharith, H. M. Alharthi, W. M. Alghamdi, Y. M. Alsenbel, N. Aslam, I. U. Khan, S. Y. Shahin, S. Dianišková, M. S. Alhareky, and K. K. Barouch, "A deep learning-based approach for the detection of early signs of gingivitis in orthodontic patients using faster region-based convolutional neural networks," *International Journal of Environmental Research and Public Health*, vol. 17, no. 22, p. 8447, 2020.

[25] M. Ezhov, M. Gusarev, M. Golitsyna, J. M. Yates, E. Kushnerev, D. Tamimi, S. Aksoy, E. Shumilov, A. Sanders, and K. Orhan, "Clinically applicable artificial intelligence system for dental diagnosis with cbct," *Scientific reports*, vol. 11, no. 1, pp. 1–16, 2021.

[26] W. You, A. Hao, S. Li, Y. Wang, and B. Xia, "Deep learning-based dental plaque detection on primary teeth: a comparison with clinical assessments," *BMC Oral Health*, vol. 20, pp. 1–7, 2020.

[27] V. L. Kouznetsova, J. Li, E. Romm, and I. F. Tsigelny, "Finding distinctions between oral cancer and periodontitis using saliva metabolites and machine learning," *Oral diseases*, vol. 27, no. 3, pp. 484–493, 2021.

[28] Y. Liang, H. W. Fan, Z. Fang, L. Miao, W. Li, X. Zhang, W. Sun, K. Wang, L. He, and X. Chen, "Oralcam: enabling self-examination and awareness of oral health using a smartphone camera," in *Proceedings of the 2020 CHI Conference on Human Factors in Computing Systems*, 2020, pp. 1–13.

[29] J.-H. Lee, D.-h. Kim, S.-N. Jeong, and S.-H. Choi, "Diagnosis and prediction of periodontally compromised teeth using a deep learning-based convolutional neural network algorithm," *Journal of periodontal & implant science*, vol. 48, no. 2, p. 114, 2018.

[30] J. Krois, T. Ekert, L. Meinhold, T. Golla, B. Kharbot, A. Wittemeier, C. Dörfer, and F. Schwendicke, "Deep learning for the radiographic detection of periodontal bone loss," *Scientific reports*, vol. 9, no. 1, pp. 1–6, 2019.

[31] J. Kim, H.-S. Lee, I.-S. Song, and K.-H. Jung, "Dentnet: Deep neural transfer network for the detection of periodontal bone loss using panoramic dental radiographs," *Scientific reports*, vol. 9, no. 1, pp. 1–9, 2019.

[32] H.-J. Chang, S.-J. Lee, T.-H. Yong, N.-Y. Shin, B.-G. Jang, J.-E. Kim, K.-H. Huh, S.-S. Lee, M.-S. Heo, S.-C. Choi *et al.*, "Deep learning hybrid method to automatically diagnose periodontal bone loss and stage periodontitis," *Scientific reports*, vol. 10, no. 1, pp. 1–8, 2020.

[33] K. Hellén-Halme, A. Lith, and X.-Q. Shi, "Reliability of marginal bone level measurements on digital panoramic and digital intraoral radiographs," *Oral radiology*, vol. 36, no. 2, pp. 135–140, 2020.

[34] B. V. Saberi, S. Nemati, M. Malekzadeh, and A. Javanmard, "Assessment of digital panoramic radiography's diagnostic value in angular bony lesions with 5 mm or deeper pocket depth in mandibular molars," *Dental research journal*, vol. 14, no. 1, p. 32, 2017.

[35] E. A. Pepelassi and A. Diamanti-Kipioti, "Selection of the most accurate method of conventional radiography for the assessment of periodontal osseous destruction," *Journal of clinical periodontology*, vol. 24, no. 8, pp. 557–567, 1997.

[36] O. Ronneberger, P. Fischer, and T. Brox, "U-net: Convolutional networks for biomedical image segmentation," in *International Conference on Medical image computing and computer-assisted intervention*. Springer, 2015, pp. 234–241.

[37] K. He, X. Zhang, S. Ren, and J. Sun, "Deep residual learning for image recognition," in *Proceedings of the IEEE conference on computer vision and pattern recognition*, 2016, pp. 770–778.

[38] L. Shen, L. R. Margolies, J. H. Rothstein, E. Fluder, R. McBride, and W. Sieh, "Deep learning to improve breast cancer detection on screening mammography," *Scientific reports*, vol. 9, no. 1, pp. 1–12, 2019.